# A study on general visual categorization of objects into animal and plant groups using global shape descriptors with a focus on category-specific deficits

**Zahra Sadeghi**

Faculty of Computer Science, Dalhousie University
zahras@dal.ca

## Abstract

*How do humans distinguish between general categories of objects? In a number of semantic-category deficits, patients are good at making broad categorization but are unable to remember fine and specific details. It has been well accepted that general information about concepts is more robust to damages related to semantic memory. Results from patients with semantic memory disorders demonstrate the loss of ability in subcategory recognition. In this paper, we review the behavioural evidence for category specific disorder and show that general categories of animal and plant are visually distinguishable without processing textural information. To this aim, we utilize shape descriptors with an additional phase of feature learning. The results are evaluated with both supervised and unsupervised learning mechanisms and confirm that the proposed method can effectively discriminates between animal and plant object categories in visual domain.*

## 1. INTRODUCTION

General categorization of items is an important topic with diverse Internet of Things (IoT) applications [1]. Researching general concepts provides high-level information about broader (i.e., superordinate or basic level) categories of objects, aiding in decisions regarding more specific subordinate categories. In many tasks, acquiring general information is essential as it streamlines further processing by offering a rapid understanding of the item in question. For instance, categorizing items into general groupings can facilitate faster searching and retrieval of relevant content. Developing general-purpose machines, known as Artificial General Intelligence (AGI), has recently gained substantial attention as it can help AI to approach human-level intelligence. In contrast to the specific-purpose systems aka "narrow AI" which are trained or fine-tuned to handle particular tasks, general intelligent machines or "strong AI" aim to be more capable of finding solutions or making decisions across various situations [2][3].
AGI can be considered as a human cognitive ability but the underlying brain mechanism for feature processing of this task is not well understood. Deep Neural Networks (DNNs) and their recent counterpart, Large Language Models (LLMs), have provided a framework for developing general purpose machines [4]. However, these deep learning-based models often struggle with semantic understanding and interpretability [5]. One way to hypothesize about features involved in general categorization is through studying cognitive disorders that impact broad categorization. Studying category-specific disorders will help understanding the inner functionality that is involved in object categorization. In this paper, we propose a feature engineering solution for designing global features that can advantage general visual categorization of living items into animal and plant groups. First, we review the category-specific disorders in the domain of cognitive science. Next, we propose a machine learning method for general object recognition and present the results. Finally, we close the paper by providing discussion and conclusion.

## 2. CATEGORY-SPECIFIC DEFICITS AND GENERAL CATEGORIZATION

Category-specific deficits are considered as one of major sources of information for understanding the recognition mechanism in brain. These impairments affect recalling information about certain category of objects. Thereupon, studying patients with naming impairment on categorization leads to important discoveries about knowledge organization in memory. It can also provide clues about concept encoding and representation in the brain.

Leveraging semantic similarities can be considered a potent strategy for modelling general categorization in a manner resembling how the human brain operates. In particular, according to the research in cognitive science, people identify objects by utilizing semantic knowledge which is stored in the part of their long-term memory called semantic memory. Semantic memory stores semantic knowledge about the world including information about category of items across different modalities and the relationships between [6]. Therefore, the study of semantic memory impairments can provide important information about object perception and organization.

According to the hierarchical structure theory, conceptual information seems to be organized in three main levels, i.e., superordinate, basic and subordinate [7]. Support for the hierarchical structure comes from neurological studies of patients with memory impairments, which show superordinate-advantage in object naming and recognition. High-level object recognition, encompassing superordinate or basic categories, is more robust, as it fosters a broad understanding that can prompt swift and appropriate responses

This impairment has highly reported in living object category rather than non-living objects [8]. The most likely explanation for this evidence is the higher amount of visual correlation between semantically related items that can be found in living group than the non-living group [9][10]. Based on the results which are obtained by behavioural studies, specific memory deficits manifest itself in object recognition and/or object identification. Semantic dementia is a well-documented semantic memory syndrome, which is caused by deterioration of semantic memory [11] and is characterized with loss of ability in recognizing subordinate categories (e.g. Horse) while the distinction between categories at basic level (e.g. Animal) of recognition is preserved for a long time. In other words, patients with semantic dementia generally fail at recognition of objects from subordinate levels, while they can perform well at more general levels or superordinate concepts (e.g., [11][12]).This syndrome is mostly related to living things as the within similarity between subordinate object of this group is relevantly high in comparison to non-living things. The living subcategories which are usually affected are animals and plants. Due to evolutionary survival instincts, recognition of living objects (i.e. animal and plant categories) is necessary for human-being [13]. Experimental results also suggest that corresponding knowledge of animal and plant concepts is organized and stored separately [14][15]. From computational standpoint this may put it forward that information regarding to each of these categories are encoded differently in brain.

Semantic memory stores visual information about the appearance of objects as well as information collected from other sensory modalities. However, experimental results suggest the higher impact of visual information in categorical knowledge representation than other modalities for living objects [16]. The role of visual inputs appears critical in knowledge representation, as damage to the visual cortex can lead to recognition difficulties. [17]. One such case is visual agnosia, a visual recognition disorder caused by damage to ventral stream pathway. This impairment is characterized by failure in visual object recognition. Specifically, patients with particular type of visual agnosia (i.e., appreciative agnosia) display very weak ability in copying line drawing objects. In fact, a person with this kind of impairment is unable to process whole parts of the seen objects or construct a coherence representation [18]. The deficit is associated only to the visual modalities as patients can normally recognize objects through information from other sensory modalities. Besides, examples of patients with simultanagnosia tend to exhibit early perceptual processing, allowing them to capture contours and global shapes, but impairing their capacity to process detailed information [19][20] which enables them to form a basic understanding of the objects [21]. According to Humphreys et al., higher degree of resemblance in terms of global shape and visual appearance can be found among living things than to items from other categories [22]. Global shape information has also shown to be favoured by children of young age for object representation [23]. These findings suggest that global shape descriptors play a key in early recognition of objects. Thereafter, visual disorders provide valuable insights into the processes underlying visual processing and human visual recognition.

In the present study, we show that basic categories within living things are distinguishable based on their global visual characteristics. We focused on the animal vs. plant categorization and leveraged shape-descriptors using project and profile information.

## 3. Method

In the following subsections we present the proposed feature representation method. We leverage shape descriptors to capture the global distinctions between broad categories of animals and plants. Global shape descriptors encode the geometric properties of objects and are capable of finding low-cost features. In this paper, we use projection and profile descriptors which can be considered as a particular case of Radon transform [24]. This method has been applied in various image analysis as well as object categorization tasks [25][26]. Our approach is tested on image category selected from Caltech 101 database, which has been widely used for object classification and categorization [27]. The list of subcategories of animal and plant objects that are used in the simulations are collected in Table 1. Our shape analysis technique is applied to the binary representations of objects. Figure 1 demonstrates samples of binary images from each category. The following subsections presents the proposed method for defining shape descriptors.

### 3.1. Projection and Profile descriptors

*Projection* of a binary image onto a line, calculates the number of on pixels in perpendicular to each partition. It can provide a compact representation about shape information of objects. We perform projection analysis of a two-dimensional binary image ($B$) onto $x$ and $y$ axes, which are known as horizontal ($H$) and vertical ($V$) projections:

$$\mathrm{H[i]} = \sum_{j=0}^{m-1} B[i,j] \qquad (1)$$

$$\mathrm{V[j]} = \sum_{i=0}^{n-1} B[i,j] \qquad (2)$$

Where, $m$ and $n$ are the number of rows and columns of an image, respectively. As mentioned earlier, all images are first rescaled to 100 by 100. Therefore, the horizontal and vertical projection descriptors are vectors of size 100.

Table 1. List of subcategories of animal and plant objects

| | |
|---|---|
| **Animals** | beaver, cugar_body, crocodile, dolphine, elephant, emu, flamingo, gerenuk, hawksbill, hedgehog, leopards, llama, okapi, pigeon, platypus, rhino, rooster, wild_cat |
| **Plants** | bonsai, joushua_tree, lotus, nautilus, strawberry, sunflower, water_lilly |

Another useful and simple shape descriptor set which is widely used in handwritten recognition is based on distance *profiles*. In this approach, distance of the outer edge of the object to the four boundary sides of image (i.e., left, right, top, and bottom) is calculated [28] [29] which results if four 100-dimensional. An example of projection and profile features for an object from animal and plant categories is illustrated in Figure 2. Projection and profile descriptors are normalized for all objects and their first four statistical features (i.e., mean, variance, skewness, and kurtosis) are measured and displayed in Figure 3 and Figure 4.

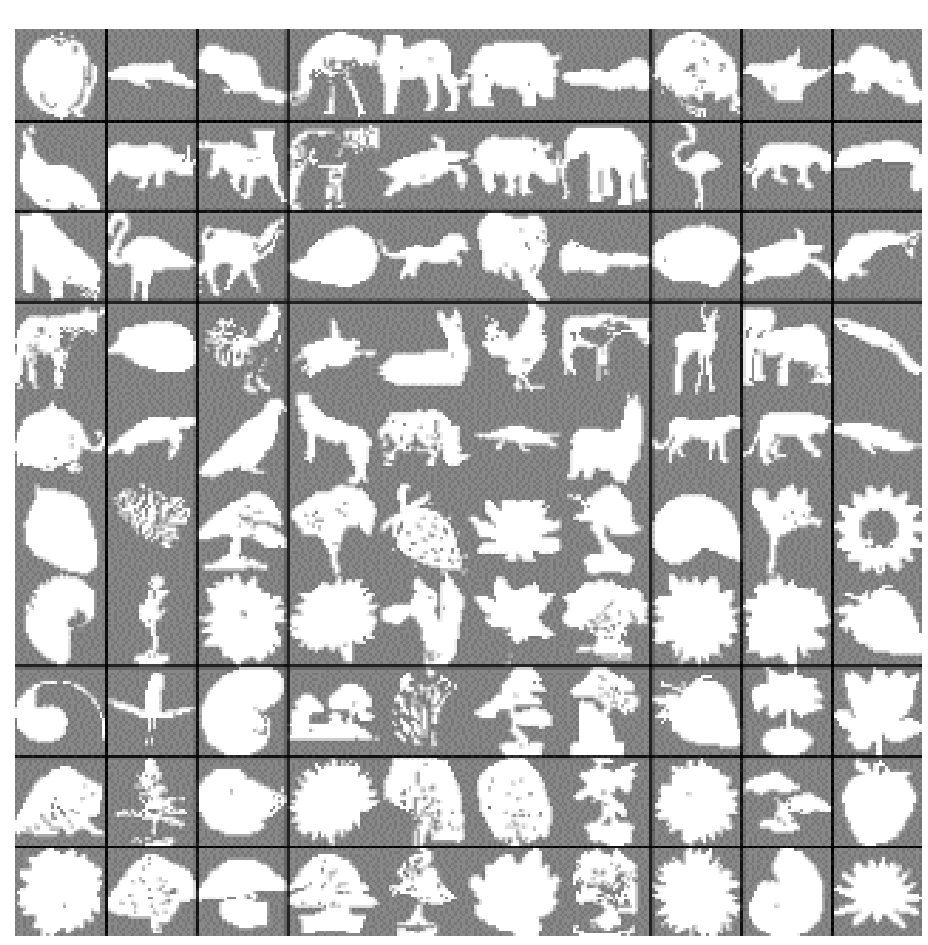

Figure 1. Sample of binary objects

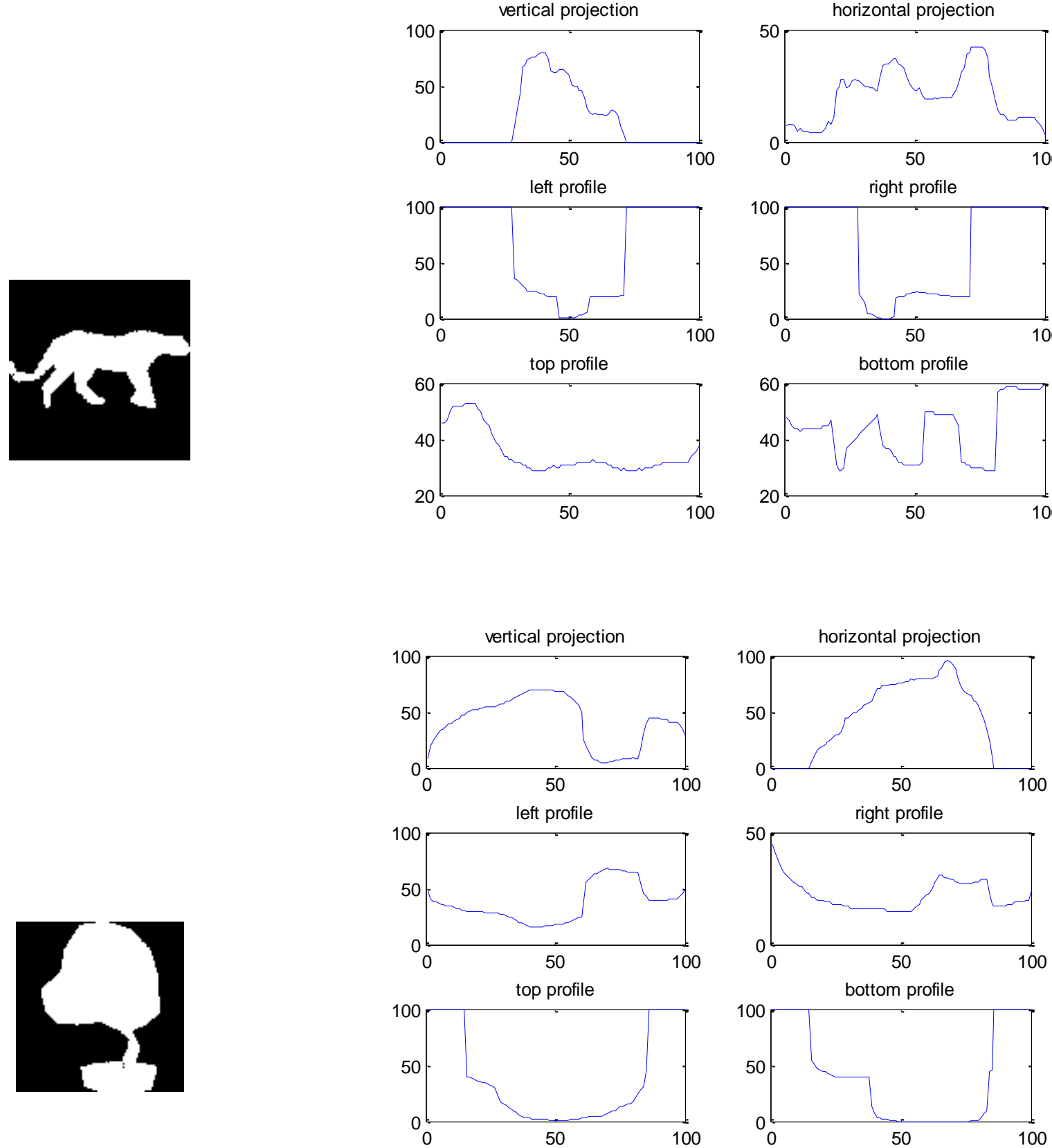

Figure 2. Example of profile and projection descriptors for an object from animal and plant categories. The x-axis shows feature array and the y-axis demonstrates the corresponding feature value.

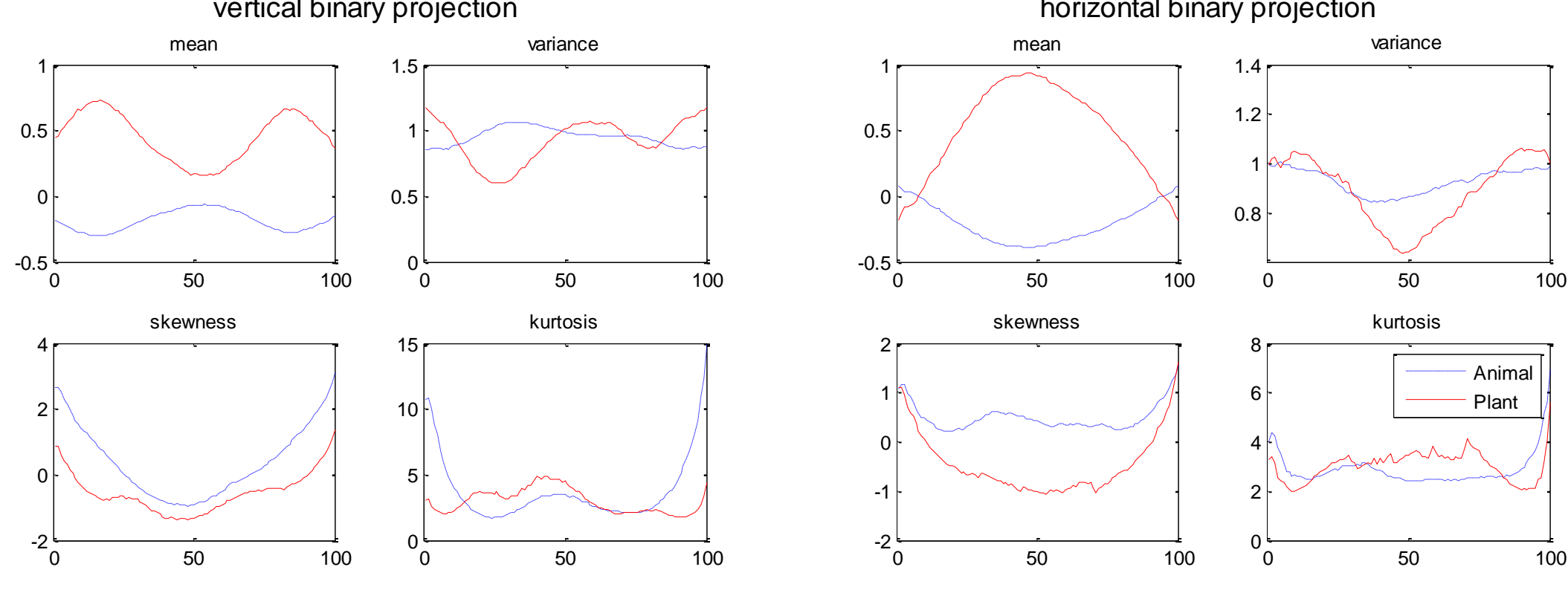

Figure 3. First four moments applied to the projection descriptors. x-axis shows feature array and y-axis demonstrates the corresponding feature value.

## 4. Evaluation

In this section, we assess the distinction capability of the proposed features to differentiate between general categories of animals and plants. To this end, we perform supervised and unsupervised algorithms. Supervised algorithms directly measure how well features contribute to classification of items, whereas unsupervised techniques groups objects based on their internal similarity characteristic. We employ SVM for supervised evaluation and k-means clustering for unsupervised evaluation, respectively. We also propose to apply a deep learning technique in order to map the features into a higher dimensional space using RBM network.

### 4.1. K-means Clustering

K-means clustering is one of the simplest and fastest unsupervised learning methods, and it has been widely adopted across a variety of applications [30]. Over time, numerous variations of this foundational algorithm have been developed [31]. Given the predetermined number of clusters, k, the k-means method can be regarded as a parameter-free algorithm. K-means clustering partitions a dataset into K clusters by iteratively assigning data points to the nearest cluster centroid and updating centroids based on the mean of assigned points, aiming to minimize the sum of squared distances within clusters. This process continues until centroids stabilize, producing K distinct clusters based on proximity to centroids in the feature space.
We perform a k-means clustering on the calculated shape descriptors to measure how much distinction can be made between animal and plant categories in an unsupervised manner. The clustering task has been applied both to the individual representation of each descriptor as well as to the concatenated vectors of those descriptors corresponding to the highest f1-score. The evaluation is repeated with 10 different random initial values and the best results are reported. The numerical performance with k-means clustering is measured using equations 3 to 6 and is presented in Figure 5. As can be implied from the results, the best distinction is obtained by vertical projection, horizontal projection and top profile.

$$P = \frac{TP}{TP+FP} \tag{3}$$

$$R = \frac{TP}{TP+FN} \tag{4}$$

$$\text{f1-score} = 2\frac{P.R}{P+R} \tag{5}$$

$$\text{Accuracy} = \frac{TP+TN}{TP+FP+FN+TN} \tag{6}$$

### 4.2. Support Vector Machine (SVM)

Supervised learning is performed by using SVM, which is a maximum margin classifier which work by finding the hyperplane that maximizes the margin between classes in the feature space and therefore best separates the classes. By using the samples on the margin (i.e., support vectors) it creates a binary model on the training data which divides the space into two separable categories. Here, we used 60% of images in train phase randomly and the rest of images are used for evaluation. SVM classifiers are applied to each of the shape-feature vectors individually. Furthermore, the concatenation of feature vectors corresponding to the best-scored accuracy is also evaluated. The mean accuracy over ten independent random splits of train and test images are listed in Figure 6. The best classification performance presents 81.4% of accuracy. These results indicate that global information derived from shape descriptors is a promising option for making a good distinction between visual images of animals and plants.

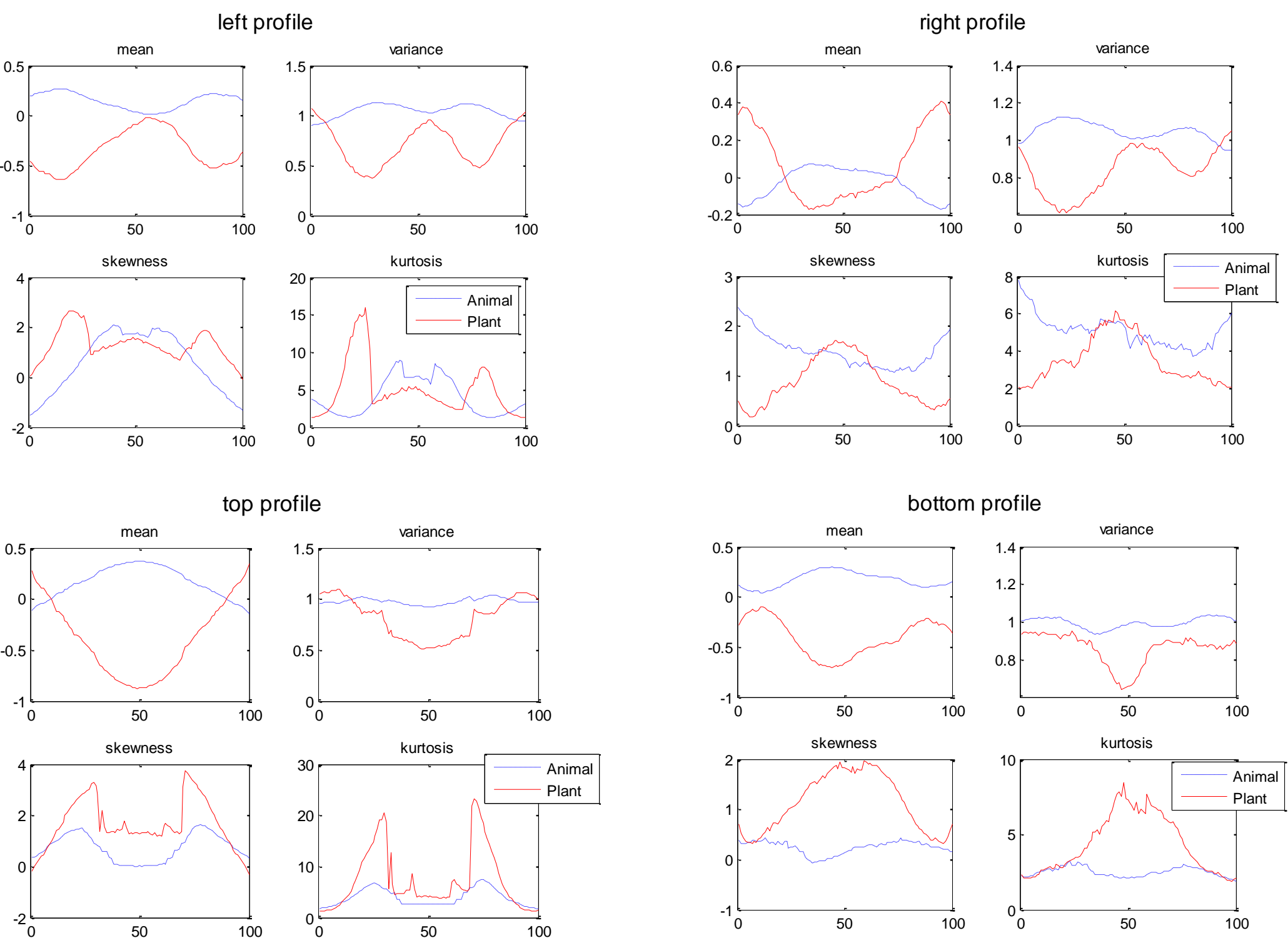

Figure 4. First four moments applied to the profile descriptors. The X-axis displays data samples, while the y-axis represents their corresponding feature values.

### 4.3. Feature learning with RBM

Restricted Boltzmann Machine (RBM) is a stochastic artificial neural network that learns features in an unsupervised manner. Technically, it consists of two layers, i.e., visible and hidden layers. Units in each layer are non-connected, but they are fully connected to the units in other layer. This model has been applied to many problems and has demonstrated high representation power [32][33][34]. We applied RBM to the calculated shape descriptors to obtain a rich representation. Training phase is performed using contrastive divergence algorithm. We trained a single-layer RBM neural network with learning rate of 0.1 on each of the previously calculated shape descriptors in order to measure the impact of neural representation. The training is performed in batch mode, with batch size of 50. The network is trained on 100 epochs. SVM classifiers are applied to the learned features by following the same procedure as explained before and finally the accuracy of classification is measured. The impact of number of hidden units is presented in Figure 7. It can be observed that as the number of hidden units increases the performance attained by the shape descriptors improves consequently. In addition, the neural representation which is obtained by employing feature learning mechanism using RBM network has enhanced classification accuracy. The best result is attributed to the vertical projection feature vector by showing 85% of accuracy with SVM classification.

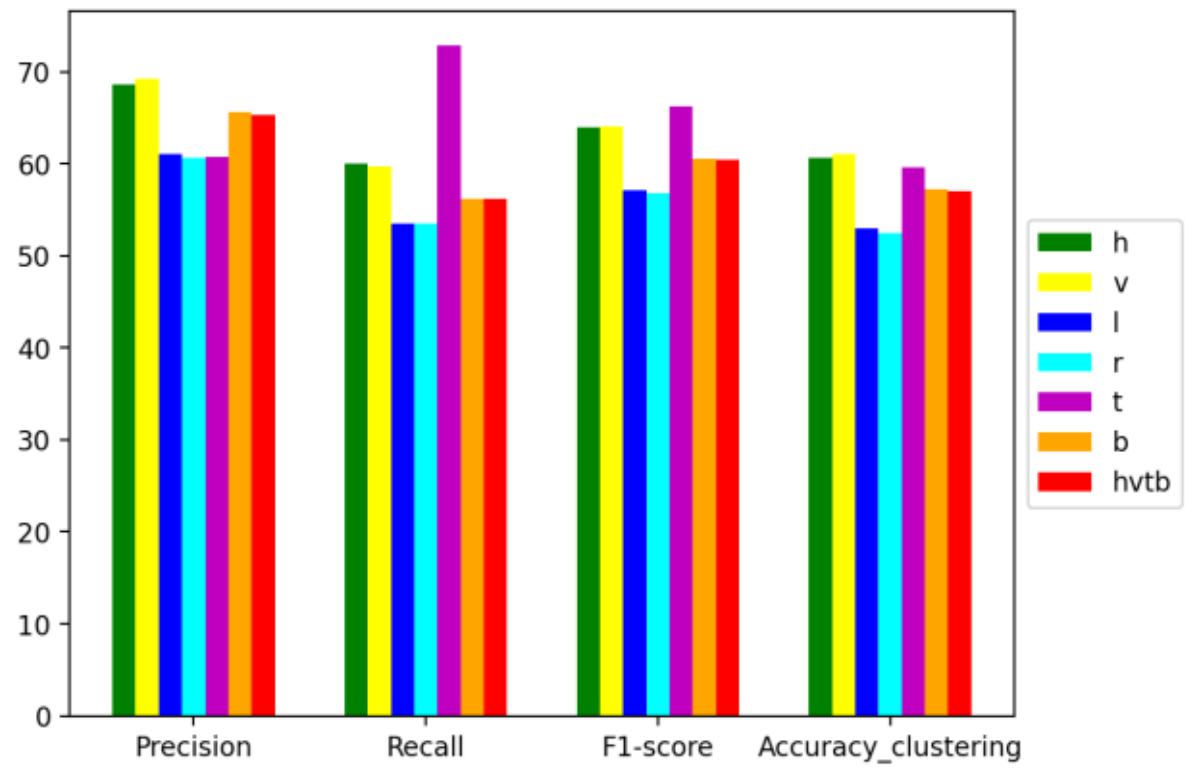

Figure 5. Clustering evaluation for input images described by h: horizontal projection, v: vertical projection, l: left profile, r: right profile, t: top profile, b: bottom profile feature vectors.

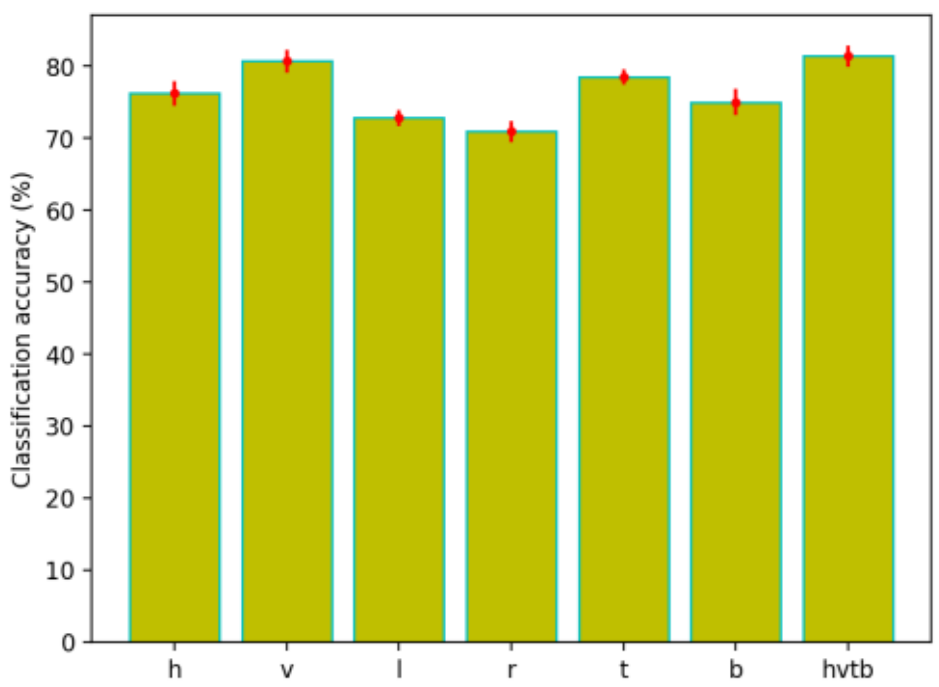

Figure 6. SVM classification accuracy for input images described by h: horizontal projection, v: vertical projection, l: left profile, r: right profile, t: top profile, b: bottom profile vectors. The results are corresponding to the average accuracy obtained over 10 independent runs with random train and test sets. Numbers in () indicate standard deviation.

## 5. Discussion and Conclusion

One question regarding the general integration is the type of required features for training models that can make general categorization. Here, we studied general categorization in the domain of visual object recognition and investigated visual features for making broad distinction between high level concepts of 'animal' and 'plant'. Experimental studies have suggested that low spatial frequency information is processed at the first glance, which leads to initial guess about the observed item [35]. The initial guesses or primitive understanding occurs at Magnocellular pathway in visual cortex [36]. Extended exposure to the viewed object enhances attention to its intricate details. This is in agreement with our result that visual projection properties, which contain low spatial frequency information, demonstrated high accuracy in both supervised and unsupervised recognition of basic categories. Furthermore, studies on children and infants' behaviour in object recognition have also demonstrated shape-based information preference for categorization. Shape information are also argued to be served as a diagnostic cue for rapid object categorization and hence is known to be decisive for the initial categorization. This preference is resulted from the fact that objects at high levels of category inclusiveness share many common features in global formation and structure. For example, animals' body basically consists of head and legs which can be considered as a distinguishable global feature. Our approach is motivated by the studies, which suggest that global information play a crucial role in making rapid and general categorization. Based on this view, we focused on the holistic structure of objects of animal and plant domains from Caltech-101 dataset[1]. We achieved precision of 70% with unsupervised learning using k-means algorithm. Moreover, our results indicate 79% accuracy in classification with the raw shape-descriptors and 85% accuracy with the learned features through RBM. In the current paper, we focused on how the two main subcategories of living things, i.e., 'animal' and 'plant' are visually presented. In some studies, 'fruit' and 'vegetable' are considered as other subcategories of living things [22]. In future studies, it will be important to assess the discriminatory power of the proposed method.

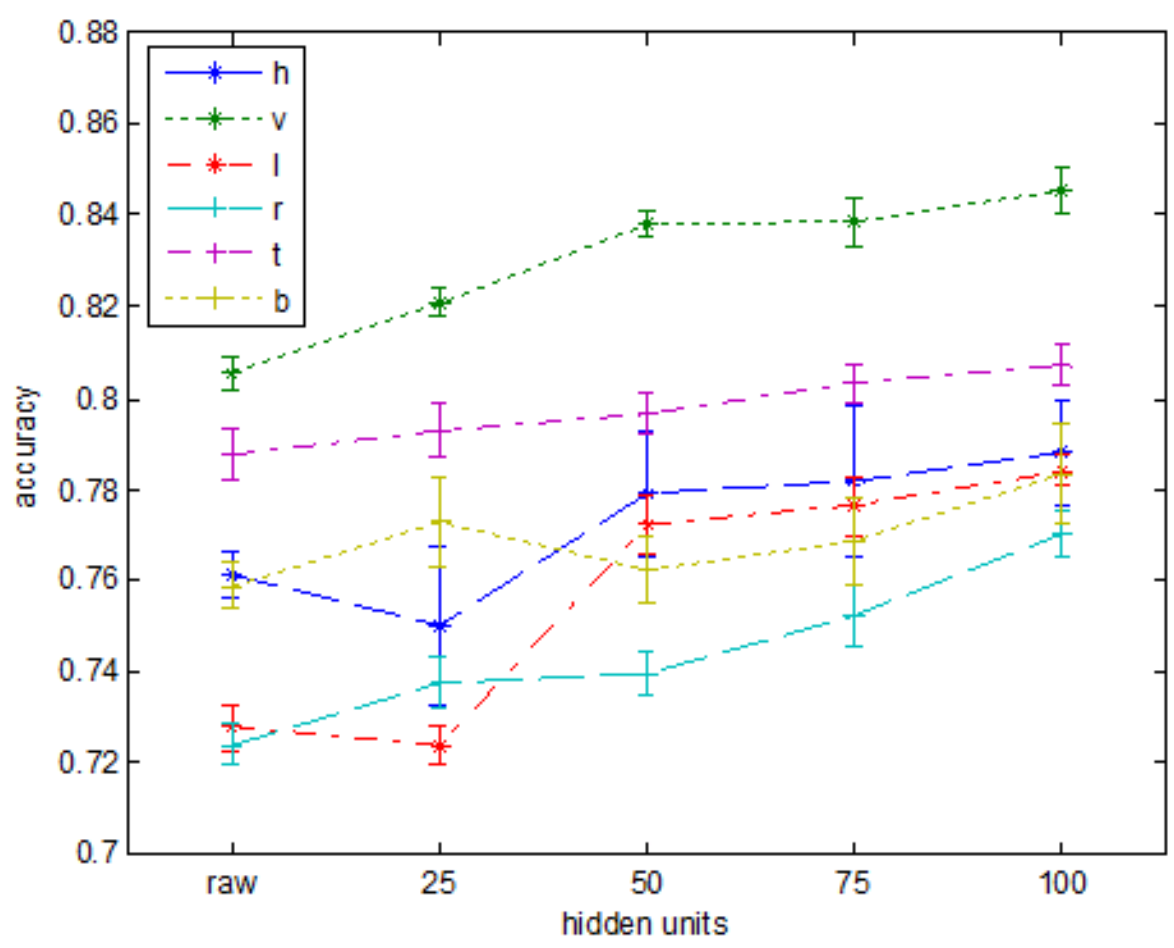

Figure 7. The effect of number of hidden units on the mean average accuracy of classification of the learned features using RBM network. The results are corresponding to 10 independent runs for each architecture. h: horizontal projection, v: vertical projection, l: left profile, r: right profile, t: top profile, b: bottom profile. Raw results correspond to the basic results without feature learning.

[1] https://data.caltech.edu/records/mzrjq-6wc02